\documentclass{amia}

\usepackage{lipsum} 
\usepackage{multirow} 
\usepackage{lscape}
\usepackage{comment}
\usepackage{xcolor}
\usepackage{tablefootnote}
\usepackage{graphicx}
\usepackage{longtable} 
\usepackage{booktabs}
\usepackage{hyperref}

\UseRawInputEncoding

\begin{document}

\title{A Scoping Review of Publicly Available Language Tasks in Clinical Natural Language Processing
}

\author{\textbf{Yanjun Gao, PhD$^1$,; Dmitriy Dligach, PhD$^2$; Leslie Christensen, MA-LIS$^3$; Samuel Tesch$^3$; Ryan Laffin$^3$; Dongfang Xu, PhD$^4$, Timothy Miller, PhD$^4$; Ozlem Uzuner, PhD$^5$;\\ Matthew M. Churpek MD, MPH, PhD$^1$; Majid Afshar, MD, MSCR$^1$ }
}

\institutes{
   \textbf{ $1$ ICU Data Science Lab, School of Medicine and Public Health, \\
    University of Wisconsin, Madison, WI \\}
    \{ygao, mchurpek, mafshar\}@medicine.wisc.edu \\ 
\textbf{$2$Department of Computer Science, Loyola University Chicago, Chicago, IL \\ }
    ddligach@luc.edu \\ 
\textbf{$3$ School of Medicine and Public Health, University of Wisconsin, Madison, WI \\ }
\{leslie.christensen, sgtesch,rlaffin\}@wisc.edu \\   
\textbf{$4$ Boston Children’s Hospital, Harvard University, Boston, MA \\}
\{Dongfang.Xu, Timothy.Miller\}@childrens.harvard.edu \\ 
\textbf{$5$ Department of Information Sciences and Technology, George Mason University,\\
Fairfax, VA }
ouzuner@gmu.edu \\ 
}

\maketitle

\section*{ABSTRACT}
\subsection*{Objective}
To provide a scoping review of papers on clinical natural language processing (NLP) tasks that use publicly available electronic health record data from a cohort of patients.

\subsection*{Materials and Methods}
 Our method followed the Preferred Reporting Items for Systematic Reviews and Meta-Analyses (PRISMA) guidelines. We searched six databases, from biomedical research (e.g. Embase (Scopus)) to computer science literature database (e.g. Association for Computational Linguistics (ACL)). A round of title/abstract screening and full-text screening were conducted by two reviewers. 

\subsection*{Results}
A total of 35 papers with 47 clinical NLP tasks met inclusion criteria between 2007 and 2021. We categorized the tasks by the type of NLP problems, including name entity recognition, entity linking, natural language inference, and other NLP tasks. Some tasks were introduced with a topic of clinical decision support applications, such as substance abuse, phenotyping, cohort selection for clinical trial. We also summarized the tasks by publication and dataset information. 

\subsection*{Discussion}
The breadth of clinical NLP tasks is still growing as the field of NLP evolves with advancements in language systems. However, gaps exist in divergent interests between general domain NLP community and clinical informatics community, and in generalizability of the data sources. We also identified issues in data selection and preparation including the lack of time-sensitive data, and invalidity of problem size and evaluation.  

\subsection*{Conclusion}
The existing clinical NLP tasks cover a wide range of topics and the field will continue to grow and attract more researchers from both general NLP domain and the clinical informatics communities. We encourage future work on proposing tasks/shared tasks that incorporate multi-disciplinary collaboration, reporting transparency, and standardization in data preparation. 

\section*{INTRODUCTION}


Since the inception of the first Integrating Biology and the Bedside (i2b2) shared task in 2006, currently known as the National Natural Language Processing (NLP) Clinical Challenge (n2c2), the field of clinical NLP has advanced in clinical applications that rely on text from the electronic health record (EHR).  Tasks with publicly available data (e.g. shared tasks) provide a new avenue for advancing the state-of-the-art using publicly available datasets in a sector that is otherwise heavily regulated and protected from sharing patient data. In an editorial approximately a decade ago, Chapman et al.\cite{chapman2011overcoming} identified the major barriers for clinical NLP developments where shared tasks may provide a solution. At the time, some of the challenges were lack of data resources including annotation tools, benchmarking and standardized metrics, reproducibility,  collaboration between the general NLP communities and health research communities, and the need for user-centered development.

Over the past decade, strides have been made with an increasing number and heterogeneity in clinical NLP tasks, and many organizers are leveraging publicly available EHR notes like the Medical Information Mart for Intensive Care (MIMC)~\cite{johnson2016mimic}. MIMIC along with clinical notes from other health systems has overcome privacy and regulatory hurdles to enable the growth of language tasks to address important clinical problems with NLP solutions. The benefits of publicly available language tasks have become apparent with an opportunity for both clinical informatics and general domain NLP communities to tackle problems together and develop systems that may translate into applied tools in health systems. The body of language tasks continues to enable the growth with complex information extraction tasks ranging from early diagnoses (e.g. substance abuse detection, phenotyping~\cite{yetisgen2017automatic,klassen2016annotating,shen2021family}) to clinical language understanding (e.g. natural language inference~\cite{abacha2019overview,romanov2018lessons}).   

However, several challenges remain as transparency in the methods, 
clinical motivation, and standardization across annotation techniques and sample size determination remain highly variable. Our objective is to review the papers on \textbf{clinical NLP tasks that use publicly available EHR data from a cohort of patients}. We aim to examine the progress over the years and describe both barriers that we have overcome as well as challenges that remain in advancing clinical NLP. This scoping review will serve as a resource for organizers and participants in the clinical NLP domain to easily retrieve details on publicly available clinical tasks as well as identify gaps and opportunities for future tasks.
As a resource to the community, we provide a listing of all the publicly available tasks from this review for easy reference\footnote{A full list can be found at \url{https://git.doit.wisc.edu/YGAO/public-available-clinical-nlp-tasks} }. 

\section*{METHODS}
The methods to conduct this scoping review adhered to standards described in the Preferred Reporting Items for Systematic Reviews and Meta-Analyses guidelines for Scoping Reviews (PRISMA-ScR).~\cite{tricco2018prisma} The review format was adapted to a population, intervention, comparator, and outcome (PICO) framework to establish the inclusion criteria for articles.~\cite{schardt2007utilization} The population consisted of a new language task for clinical NLP using publicly available EHRs from a cohort of patients. The intervention includes expert annotations to build a labelled corpus of data.  The comparator is a test dataset with an evaluation metric, and the outcome is a challenge or published model that represents a state-of-the-art (SOTA) for the task. 

\subsection*{Literature Search}
In adherence with the PICO framework, the librarian (LC) performed a full, systematic review of the literature between January 1985 and September 2021. The search combined controlled vocabulary and title/abstract terms related to the shared language tasks in clinical NLP with a focus on publicly available datasets. The search was developed in PubMed, tested against a set of exemplar articles, and then translated into the following databases: (1) Embase (Scopus); (2) The Association for Computing Machinery (ACM) Guide to Computing Literature (ACM Digital Library); (3) Science Citation Index Expanded (Web of Science); (4) Conference Proceedings Citation Index-Science (Web of Science); and (5) Emerging Sources Citation Index (Web of Science). The metadata from the Association for Computational Linguistics (ACL) Anthology was downloaded separately and searched based on the database search strategies. The search strategies were peer reviewed by two University of Wisconsin (UW)-Madison Science and Engineering librarians. All searches were performed on September 8, 2021 except for ACL, which was on September 1, 2021. No publication type, language, or date filters were applied. Results were downloaded to a citation management software (EndNote x9, Clarivate Analytics, Philadelphia, PA) and underwent manual de-duplication by the librarian. Unique records were uploaded to Rayyan screening platform~\cite{ouzzani2016rayyan} for independent review.  The full query with search terms and Boolean operations for each database is detailed in Appendix A. 

\subsection*{Study Selection}
Study selection criteria were established a priori and included the following: (1) publicly available train/test dataset from a cohort of patients; (2) NLP task; (3) novel benchmark metric; (4) models that were built and tested to establish state-of-the-art results for the novel benchmark metric; and (5) English-language research articles and tasks. Articles were excluded if the tasks were focused on the biomedical domain (genomics data, non-patient data, data from clinical research databases including PubMed articles), subject-matter specific tasks without publicly available data, pre-prints or non peer-reviewed, and individual use-case systems not designed as a shared task. Multiple papers shared a data challenge with multiple tracks. For example, the 2014 i2b2/UTHealth shared task had two tracks, protected health information (PHI) de-identification and temporal identification of risk factors for heart disease. We analyzed each track as its own task, and some tasks consisted of multiple subtasks. If the subtask focused on a distinct clinical problem, then we also considered each subtask as its own task. We excluded subtasks when the data were not clinical text and it was not related to clinical NLP. 

Review of titles and abstracts for inclusion into full-text article review was performed by researchers with expertise in NLP and clinical informatics (YG and MA). The first 400 titles/abstracts were reviewed by the two reviewers in a blinded fashion and a Cohen’s Kappa score for inter-annotator agreement was above 0.80. The subsequent papers were divided and reviewed by each reviewer independently. Any disagreements or indeterminate decisions were resolved through discussion and consensus.

\subsection*{Data Synthesis and Summarization}
Among the papers included in the scoping review, characteristics of the shared tasks were described and the data corpus metrics were summarized into Tables.  The following characteristics were provided: (1) publication date and location; (2) the type of NLP task and data source; (3) level of annotation; (4) participant details; (5) data corpus details; and (6) evaluation metrics. Depending on where the task was published, we categorized each task as originating from the Clinical Informatics or general domain NLP community. No critical appraisal of the literature was performed because of the heterogeneity in tasks and evaluation metrics. We followed the guideline and checklist from the 2018 PRISMA-ScR (Appendix B).\cite{tricco2018prisma}

\section*{RESULTS}


\subsection*{Search Results}
\begin{figure}
    \centering
    \includegraphics[scale=0.8]{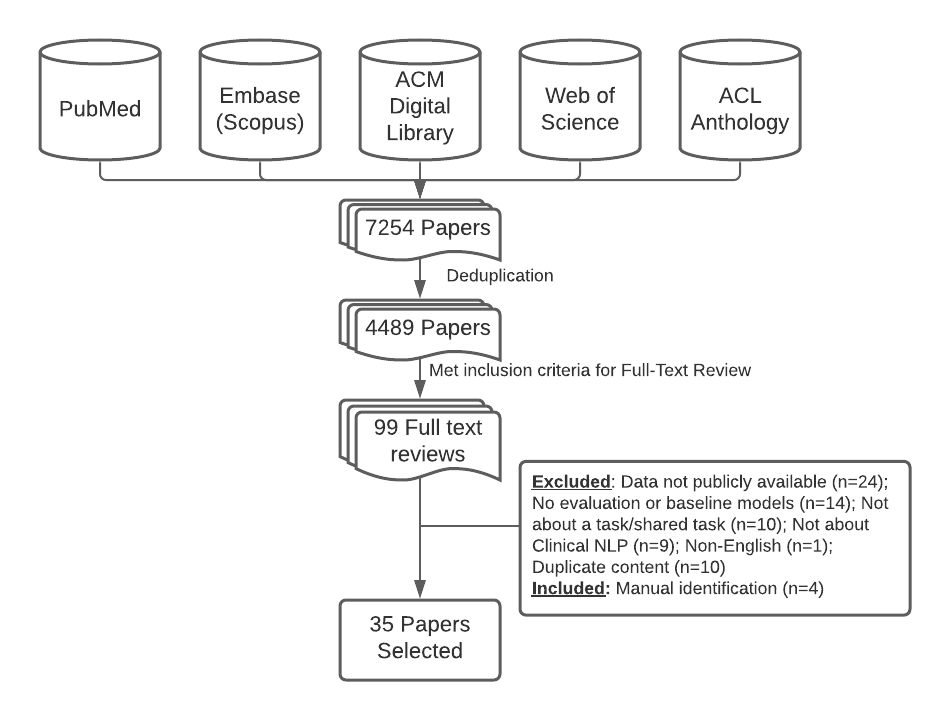}
    \caption{ PRISMA Diagram of our paper review process. }
    \label{fig:review}
\end{figure}

Our search results identified 4,489 abstracts for review after de-duplication. After the first review phase with title/abstract screening, 99 papers met inclusion criteria for full-text review. During the full-text review phase, 68 papers were excluded and the most common reason for exclusion was not having publicly available data (n=24). During the full-text review, another five papers were identified that were not part of the original query results. Thirty-five papers spanning 47 clinical NLP tasks between 2007 and 2021 were ultimately included for analysis. Figure~\ref{fig:review} illustrates the selection process and results. All of the included papers were published in peer-reviewed clinical informatics (CI) and general domain NLP journals and conference proceedings.

\subsection*{General Characteristics of Included Papers}
\begin{figure}
    \centering
    \includegraphics[scale=0.40]{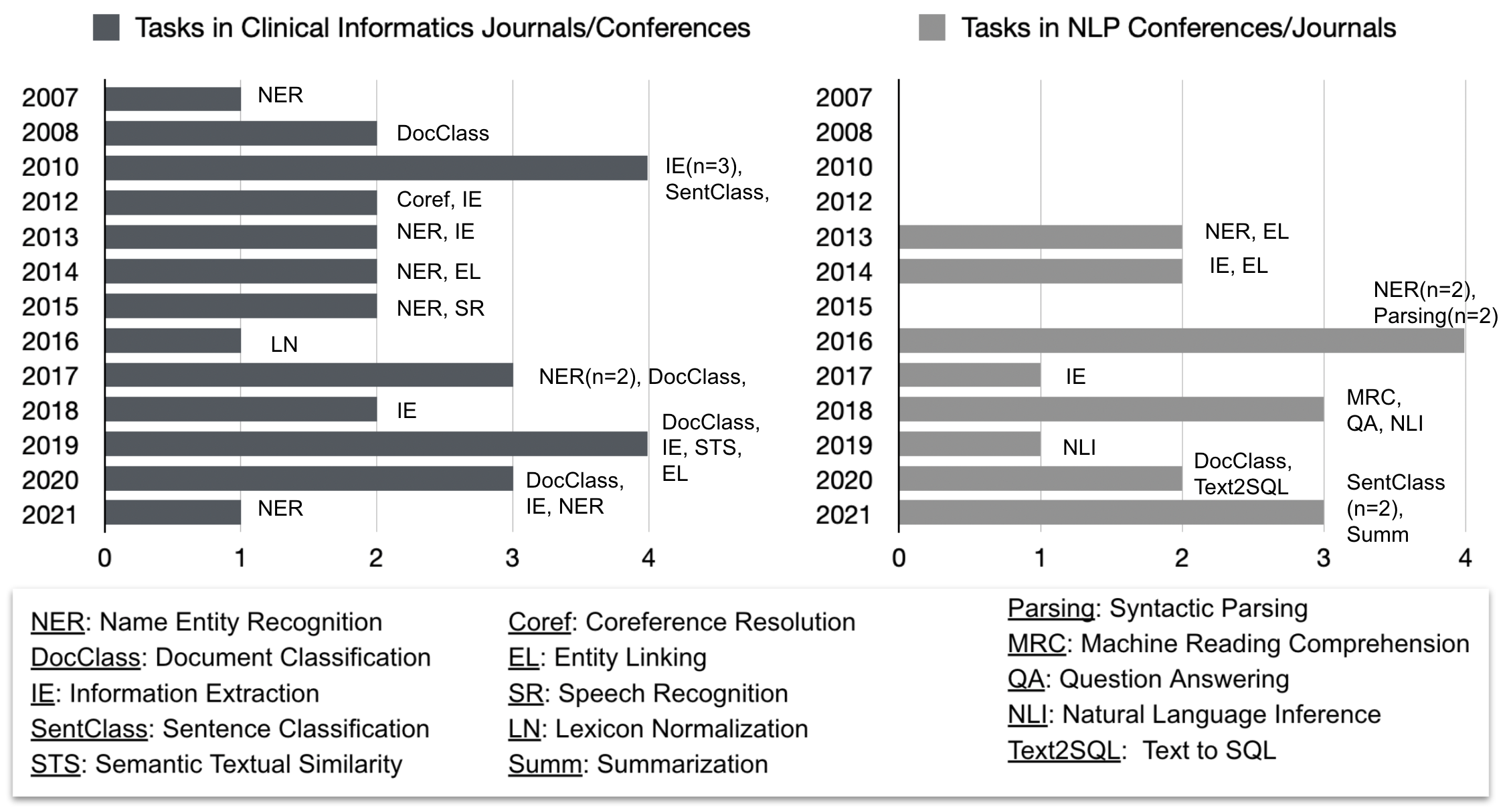}
    \caption{Types of tasks and community interests across years. } 
    \label{fig:interest}
\end{figure}


The majority of tasks appeared in CI journals with the most frequently occurring in the Journal of the American Medical Informatics Association (JAMIA; n=11)~\cite{uzuner2007evaluating,uzuner2008identifying,uzuner2009recognizing,uzuner2010extracting,uzuner2011i2b2va,uzuner2012evaluating,sun2013evaluating,pradhan2015evaluating,henry2020n2cc,henry20202018,stubbs2019cohort}, the Journal of Biomedical Informatics (JBI; n=5)~\cite{stubbs2015identifying,stubbs2015a,filannino2017symptom,stubbs2017deidentification,lybarger2021annotating}, and the Journal of Medical Informatics (JMIR; n=2)~\cite{wang2020n2c2,shen2021family}. The remaining papers were distributed across other health/clinical informatics journals and conference proceedings, including Artificial Intelligence in Medicine~\cite{yetisgen2017automatic}, Journal of Biomedical Semantics~\cite{mowery2016normalizing}, Drug Safety~\cite{jagannatha2019overview}, and conference proceedings from American Medical Informatics Association Symposium (AMIA, n=2)~\cite{uzuner2008second,peng2018negbio} and World Congress on Medical and Health Informatics (MEDINFO)~\cite{viani2019annotating}. In the general domain NLP community, the major proceedings included the Association of Computational Linguistics (ACL, n=2)~\cite{mullenbach-etal-2021-clip,yue2020clinical}, International Conference on Language Resources and Evaluation (LREC, n=2)~\cite{klassen2016annotating,moseley-etal-2020-corpus}, and Empirical Methods in Natural Language Processing (EMNLP, n=2)~\cite{pampari2018emrqa,romanov2018lessons}, International Conference of the Cross-Language Evaluation Forum for European Language (or known as Conferences and Labs of Evaluation Forum, CLEF, n=3)~\cite{kelly2014overview,suominen2013overview,goeuriot2015overview}, and International World Wide Web Conference (WWW)~\cite{wang2020text}. Some tasks were also published in workshops such as the International Workshop on Semantic Evaluation (SemEval, n=2)~\cite{pradhan2014semeval,bethard2017semeval}, Biomedical Natural Language Processing Workshop (BioNLP, n=2)~\cite{abacha2019overview,abacha2021overview}, and Workshop on Natural Language Processing for Medical Conversation~\cite{van2021assertion}. One paper publishes in the journal of Language Resources and Evaluation~\cite{savkov2016annotating}.

The peer review process and target audiences between CI and NLP publications are considerably different. For this reason, where the task is published indicates an interest rising from CI and the general NLP communities. 
Several differences exist in the types of tasks shared between these two communities. Figure~\ref{fig:interest} illustrates the type of tasks and counts between 2007 and 2021 from the two communities. Overall, 28 of the tasks were published by the CI community, and 19 tasks were published by the general domain NLP community. The earliest shared task was published in a CI journal in 2007 (i2b2 Protected Health Information (PHI) De-Identification~\cite{uzuner2007evaluating}), showing a longer history in developing clinical NLP tasks among the CI community.  Six years later, the NLP community published its first clinical NLP task in the 2013 CLEF eHealth Task 2 Disorder Mention~\cite{suominen2013overview}. Interests from the general domain NLP community have been increasing across the years, representing the majority of shared tasks in 2021 (Summarization~\cite{abacha2021overview}, Action Item Extraction~\cite{mullenbach-etal-2021-clip}, Assertion Detection~\cite{van2021assertion}). 

Name Entity Recognition (NER) represents nearly a quarter of all tasks (23.40\%, n=11) with 28.57\% (n=8) and 15.79\% (n=3) in CI~\cite{jagannatha2019overview,shen2021family,stubbs2015identifying,uzuner2007evaluating, stubbs2015a,stubbs2017deidentification,yetisgen2017automatic,henry20202018} and general domain NLP papers~\cite{suominen2013overview,savkov2016annotating,klassen2016annotating}, respectively. Other tasks that occurred frequently in CI were Information Extraction (IE; n=7)~\cite{kelly2014overview,sun2013evaluating,bethard2017semeval,uzuner2011i2b2va,viani2019annotating,henry20202018}, Document Classification (DocClass; n=5)~\cite{filannino2017symptom,lybarger2021annotating,stubbs2019cohort,uzuner2009recognizing,uzuner2008identifying}. In the general domain NLP community, the types of tasks were distributed relatively evenly across Entity Linking (EL; n=2)~\cite{suominen2013overview,pradhan2014semeval}, Syntactic Parsing (Parsing; n=2)~\cite{klassen2016annotating,savkov2016annotating}, Natural Language Inference (NLI; n=2)~\cite{romanov2018lessons,abacha2019overview}, and Sentence Classification (SentClass; n=2)~\cite{mullenbach-etal-2021-clip,peng2018negbio}. Tasks required text understanding and generation are proposed by general NLP community, such as Machine Reading Comprehension (MRC)~\cite{yue2020clinical}, Summarization (Summ)~\cite{abacha2021overview} and Question Answering (QA)~\cite{pampari2018emrqa}. 


\subsection*{Descriptions of Included Tasks and Data}

The characteristics of the tasks are shown in Table~\ref{tab1:datasets}. We found that 38\% of the NLP tasks are introduced with an impact on clinical decision making. Most of the clinical impacts are under the category of NER task, introducing detection and identification of various medical conditions~\cite{pradhan2015evaluating}, substance abuse~\cite{yetisgen2017automatic}, medical risk factors~\cite{stubbs2015identifying}, medical events~\cite{sun2013evaluating,henry20202018}, and PHI de-identification~\cite{uzuner2007evaluating, stubbs2015a,stubbs2017deidentification}.  Phenotyping \cite{savkov2016annotating} introduced a corpus annotated with NER without identifying a specific clinical purpose. DocClass is the second most frequent NLP task after NER (12.76\%, n=6), covering clinical decision support applications from symptom severity~\cite{filannino2017symptom}, obesity and comorbidity phenotyping~\cite{uzuner2009recognizing}, smoking status~\cite{uzuner2008identifying}, to cohort selection for clinical trials~\cite{stubbs2019cohort}. Inconsistencies in defining NLP tasks occurred with two papers~\cite{klassen2016annotating,shen2021family} 
that described phenotyping as an NER task and others~\cite{moseley-etal-2020-corpus,lybarger2021annotating} 
described it as a document classification task. Tasks without specific clinical applications were NLI~\cite{romanov2018lessons,abacha2019overview}, MRC~\cite{yue2020clinical}, QA~\cite{pampari2018emrqa}, Summ~\cite{abacha2021overview}, STS~\cite{wang2020n2c2}, Coref~\cite{uzuner2012evaluating}, Parsing~\cite{klassen2016annotating,savkov2016annotating}, Text2SQL~\cite{wang2020text} and SR~\cite{goeuriot2015overview}. Most of these tasks were introduced by the general domain NLP community, except STS~\cite{wang2020n2c2} and Coref~\cite{uzuner2012evaluating}. 

The data sources used to build the corpora were frequently derived from single health systems.  Among them, the most frequent was from MIMIC~\cite{johnson2016mimic}, which is from a large tertiary academic center in Boston and represented 31.91\% (n=15) of the tasks~\cite{abacha2021overview,mullenbach-etal-2021-clip,van2021assertion,wang2020text,lybarger2021annotating,moseley-etal-2020-corpus,abacha2019overview,henry20202018,romanov2018lessons,mowery2016normalizing,pradhan2014semeval,suominen2013overview,kelly2014overview}. Other urban and academic health systems also contributed by releasing their data in a de-identified format including the following: Partners HealthCare (PHC, n=8)~\cite{henry2020n2cc,yue2020clinical,stubbs2017deidentification,filannino2017symptom,sun2013evaluating,uzuner2010extracting,uzuner2008identifying,uzuner2009recognizing,uzuner2007evaluating}; Beth Israel Deaconess Medical Center (BIDMC, n=7)~\cite{uzuner2011i2b2va,sun2013evaluating,henry2020n2cc,yue2020clinical}; University of Pittsburgh Medical Center (UPMC, n=5)~\cite{yue2020clinical,uzuner2012evaluating,uzuner2011i2b2va}; University of Texas Health System (UTHealth, n=4)~\cite{uzuner2011i2b2va,uzuner2012evaluating,yue2020clinical}; Mayo Clinic (Mayo, n=3)~\cite{bethard2017semeval,wang2020n2c2,shen2021family}; and University of Washington Harborview Medical Center (UW Harborview, n=3)~\cite{klassen2016annotating,lybarger2021annotating}. All these data sources represent single centers that are tertiary academic medical centers.  

Several papers remain general in describing the note types as ``EMR" or ``EHR" without further specifying the type of note (e.g., progress note, discharge summary, radiology report, etc.). For those papers, we denoted the type as ``clinical notes" (n=14)~\cite{stubbs2015a,pradhan2014semeval,stubbs2015identifying,savkov2016annotating,stubbs2017deidentification,stubbs2019cohort,wang2020n2c2,yue2020clinical,bethard2017semeval,viani2019annotating,jagannatha2019overview}.  Other papers gave clear specifications and discharge summaries were the most frequent note type (36.71\%. n=17)~\cite{mullenbach-etal-2021-clip,van2021assertion,lybarger2021annotating,moseley-etal-2020-corpus,henry2020n2cc,henry20202018,mowery2016normalizing,suominen2013overview,kelly2014overview,sun2013evaluating,uzuner2012evaluating,uzuner2011i2b2va,uzuner2010extracting,uzuner2008identifying,uzuner2009recognizing,uzuner2007evaluating}, followed by radiology reports (14.89\%, n=7)~\cite{kelly2014overview,mowery2016normalizing,peng2018negbio,abacha2021overview}. Other note types included history and physical admission (H\&P), daily progress notes, electrocardiogram, echocardiogram, pathology reports, and psychiatric evaluation records~\cite{kelly2014overview,mowery2016normalizing,filannino2017symptom,romanov2018lessons,moseley-etal-2020-corpus,shen2021family}. Text2SQL and SR had note types that were different from all other tasks. Text2SQL included structured data because their goal was to convert the tabular data into SQL language~\cite{wang2020text}. SR included audio records from nursing handover sessions with the intent of developing written text from spoken language~\cite{goeuriot2015overview}.    

We found that more than a half of the tasks used data annotated at the lexical level (51.06\%, n=24) ~\cite{suominen2013overview, henry2020n2cc,pradhan2014semeval,mowery2016normalizing,shen2021family,pradhan2015evaluating,stubbs2015identifying,uzuner2007evaluating, stubbs2015a,stubbs2017deidentification,yetisgen2017automatic,henry20202018,savkov2016annotating,klassen2016annotating,uzuner2012evaluating}. For these tasks, lexical features served as the basis for assigning labels.  
Tasks like NLI~\cite{romanov2018lessons,abacha2019overview}, STS~\cite{wang2020n2c2}, Parsing~\cite{klassen2016annotating,savkov2016annotating} and SentClass~\cite{mullenbach-etal-2021-clip,uzuner2011i2b2va,van2021assertion,peng2018negbio} had annotations at the sentence level. Document level annotation were created for tasks in DocClass~\cite{filannino2017symptom,moseley-etal-2020-corpus,lybarger2021annotating,stubbs2019cohort,uzuner2009recognizing,uzuner2008identifying}, MRC~\cite{yue2020clinical}, QA~\cite{pampari2018emrqa} and Summ~\cite{abacha2021overview}.  

\subsection*{Descriptions of Task Participation, Data Size, and Evaluation} 

Details on participants in the shared tasks along with data metrics are shown in Table~\ref{tab2:tasks}; however, some papers did not report the participants. Among all tasks where participation information is available, the number of participants ranged from 5 teams to 35 teams. Summ was only hosted once but published in 2021 as the task with  the greatest number of teams (n=35) submitting their systems~\cite{abacha2021overview}. STS, published in 2020, is the second most popular task and attracted 33 teams~\cite{wang2020n2c2}. DocClass ranked the third in number of team participants, with an average of 29.33 teams across three shared tasks~\cite{uzuner2009recognizing,uzuner2008identifying,filannino2017symptom,stubbs2019cohort}. 

Sample sizes across the labels were  highly variable and ranged from a few hundred manually annotated labels to semi-automated methods that produced several-fold more labels. None of the papers justified their sample size and they were simply reported as convenience samples. Further, not all tasks reported their data splits in the papers~\cite{sun2013evaluating,klassen2016annotating,peng2018negbio}. For example, we only found one SentClass task with data split information readily available~\cite{mullenbach-etal-2021-clip}. The units of dataset size were also heterogeneous, and sometimes not consistent with the annotation. For instance, annotation for EL tasks were created at lexicon level, yet few papers reported the size regarding number of words and tags~\cite{henry2020n2cc}. The biggest corpora task was QA, using the emrQA dataset~\cite{pampari2018emrqa}. Their annotation was generated semi-automatically on all i2b2 data. The emrQA was also used in an MRC task with a subtle difference in data split~\cite{yue2020clinical}. NLI also had large sets of sentence pairs, ranging between 11,000 and 14,000 for the train set, and 405 to 1,400 for the test set~\cite{romanov2018lessons,abacha2019overview}.   

Accuracy and F1 were the two most frequent evaluation metrics. Most tasks used one of the two metrics focused on evaluating if predicted labels were correct against a gold standard, such as EL, NER, etc. Most tasks in DocClass applied the F1 score to evaluate the predicted labels, with the exception of \cite{filannino2017symptom} which reported the Mean Absolute Error. Some tasks used metrics that were more common in general NLP domain, such as ROUGE~\cite{lin2004rouge} and BERTScore~\cite{zhang2019bertscore} for summarization evaluation~\cite{abacha2021overview}; Pearson Correlation for STS task~\cite{wang2020n2c2}. Tasks in parsing used F1 as well as Unlabeled Attachment Score (UAS) and Labeled Attachment Score (LAS), two metrics that evaluated if the predicted parsing tags were correct against human-generated parsed labels~\cite{klassen2016annotating,savkov2016annotating}.    

\hfill \break
\begin{small}
\begin{longtable}{|l|l|l|l|r|} 

\caption{\small Overview of tasks and datasets, including the data source and types, and annotation units \label{tab1:datasets}} \\ \hline 

\multirow{2}{*}{\textbf{NLP Task}} & \textbf{Impact on Clinical} & \textbf{Data } & \textbf{Note} & \textbf{Annotation} \\
 & \textbf{Decision Making} & \textbf{Source} & \textbf{Type} & \textbf{Unit} \\ 
\hline
Entity  & General Non-specific  & MIMIC(n=3), VA Salt   & discharge summaries,  & Lexicon  \\ 
Linking &   ~\cite{suominen2013overview, henry2020n2cc,pradhan2014semeval,pradhan2015evaluating} & Lake City Health & progress notes & \\ 
\hline 
Natural Language  & General Non-specific~\cite{romanov2018lessons,abacha2019overview}  & MIMIC(n=2) & history and  & Sentence \\ 
Inference & & & physical admission & Pairs \\  
\hline
Lexical Normalization & General Non-specific & MIMIC & discharge summaries, & Lexicon  \\ 
& ~\cite{mowery2016normalizing} & & electrocardiogram  & \\
& & & echocardiogram, & \\
& & & radiology report &  \\  
\hline 
Machine Reading  &  General Non-specific~\cite{yue2020clinical} 
&  PHC\footnote{Partners HealthCare, rebranded as MassGeneralBrigham }, UPMC\footnote{University of Pittsburgh Medical Center},  &  clinical notes\footnote{We use ``clinical notes" to mark the data type when the paper did not specify the note type. } & Document \\ 
Comprehension & & UTHealth, BIDMC\footnote{Beth Israel Deaconess Medical Center} &  & \\ 
\hline 
Question Answering  &  General Non-specific~\cite{pampari2018emrqa} &  PHC, BIDMC, MIMIC &  clinical notes & Document \\ 
\hline 

Summarization & General Non-specific~\cite{abacha2021overview} &  MIMIC  & radiology reports & Document  \\ 
\hline 

Name Entity  & Phenotyping~\cite{klassen2016annotating,shen2021family} & UMass Memorial  & discharge summaries & Lexicon\\
Recognition & Disorder Detection~\cite{suominen2013overview,pradhan2015evaluating} & Health, PHC(n=2), &(n=2), radiology  & \\  
&Risk Factor Identification~\cite{stubbs2015identifying}, &MTSample, BIDMC,& reports, history and & \\ 
& PHI De-identification~\cite{uzuner2007evaluating, stubbs2015a,stubbs2017deidentification} & MIMIC, UW & physical admission, & \\ 
&  Substance Abuse Detection & Harboview, Mayo\footnote{Mayo Clinic}, & clinical notes (n=6) & \\
& ~\cite{yetisgen2017automatic} & UK National Health&  &\\ 
& General Non-specific~\cite{savkov2016annotating,sun2013evaluating,henry20202018} &UTHealth(n=2) & &\\ 
\hline 
Information Extraction &  Disorder Mention~\cite{kelly2014overview} &PHC(n=2), BIDMC(n=2) & discharge summaries & Lexicon \\ 
 & General Non-specific & UPMC(n=2), MIMIC & (n=6), radiology & \\
 & (Time)~\cite{sun2013evaluating,bethard2017semeval,uzuner2011i2b2va,viani2019annotating} & (n=2), Mayo, &  report, electrocar- & \\ 
 & (Concept)~\cite{henry20202018,uzuner2011i2b2va} & UK National Health & diogram, clinical notes &\\
\hline 

Semantic Textual  & General Non-specific~\cite{wang2020n2c2} &Mayo & clinical notes &Sentence \\
Similarity & & &  & Pairs \\ 
\hline 

Co-reference Resolution & General Non-specific~\cite{uzuner2012evaluating} & PHC, BIDMC,  & discharge summaries & Lexicon \\ 
& & UPMC & & \\ 
\hline 

Syntactic Parsing & General Non-specific & UK National Health, & radiology report,  & Sentence\\
& ~\cite{klassen2016annotating,savkov2016annotating}  &UW HarborView &clinical notes & \\
\hline 
Sentence &  Action Item Extraction~\cite{mullenbach-etal-2021-clip} &  MIMIC(n=2), &discharge summaries & Sentence \\
Classification & General Non-specific & Source Not Specified &(n=2), & \\
& (Assertion)~\cite{uzuner2011i2b2va,van2021assertion} & & & \\ 
& (Negation)~\cite{peng2018negbio} & & & \\ 
\hline 

Document  & Symptom Severity & PHC(n=3), MIMIC & discharge summaries, & Document \\ 
Classification & Prediction~\cite{filannino2017symptom} & (n=2), UTHealth, & (n=4), progress &\\ 
 & Phenotyping~\cite{moseley-etal-2020-corpus,lybarger2021annotating} & UW Harborview, & notes, clinical notes, & \\ 
 & Cohort Selection for Clinical  & Harvard Medical &psychiatric evaluation  & \\ 
 & Trial~\cite{stubbs2019cohort} &School & records & \\ 
 & Obesity Classification~\cite{uzuner2009recognizing} & &  & \\ 
 & Smoking Status & &  & \\ 
 & Classification~\cite{uzuner2008identifying} & & & \\ 
\hline 

Others (Text2SQL,  & General Non-Specific & MIMIC, NCTA\footnote{Nursing Care Team Assistant} & nursing handover& Various \\
 Speech Recognition) & ~\cite{wang2020text,goeuriot2015overview} 
 & & data, structured & \\ 
 &  & &  data in EHR & \\ 
  \hline 
\end{longtable}

\end{small} 



\begin{small} 
\begin{longtable}{|l|l|l|l|l|r|} 
\caption{\small Overview of tasks, average number of participants across years, years range for publications and evaluation metrics \label{tab2:tasks}} \\ \hline 
\multirow{2}{*}{\textbf{NLP Task}} & \textbf{Avg. Number of } & \textbf{Publication  } & \multicolumn{2}{c|}{\textbf{Data Split Range}}& \textbf{Evaluation } \\
 & \textbf{Participants\footnote{We count the number of participants on team-basis with $n$ representing the number of papers when the task is a shared task or open data challenge.}} & \textbf{Years\footnote{We report the range of years for the included papers}} & \multicolumn{1}{c}{\textbf{Training}} &  \multicolumn{1}{c|}{\textbf{Test}} & \textbf{Metric} \\ 
\hline
Entity Linking  & 25.33 (n=4) & 2014-2020 & 50-199 Notes & 50-133 Notes & Acc.  \\ 
\hline 
Natural Language  & 17 (n=1) &  2018-2019 & 11k-14k   & 405-1.4k  & Acc. \\ 
Inference & & &Pairs & Pairs  & \\  \hline
Lexical Normalization & 5 (n=1) & 2016 & 199 Notes & 99 Notes & Acc. \\   \hline 
Machine Reading  &  NA & 2020 & 91k  & 9.9k  & Exact Match, F1 \\ 
Comprehension & & & Queries & Queries& \\ \hline 
Question Answering  & NA & 2018-2021 & 658k-1M Pairs & 188k-296K Pairs & Acc. \\
\hline 
Summarization & 35 (n=1) & 2021 &91k &600 &  ROUGE, HOLMS, \\ 
& & & Notes &Notes &  BERTScore, CheXBert \\ \hline 
Name Entity  & 15.57 (n=7) & 2007-2021 & 99-3.1k& 117-896 & F1, Acc.  \\
Recognition &  &  & Notes &Notes &  \\  
 \hline 
Information Extraction & 17.43 (n=7) & 2011-2019 & 300-876 Notes & 100-574 Notes & Acc., F1 \\ 
\hline 
Semantic Textual  & 33 (n=1)& 2020 &1.6k Pairs & 412 Pairs &  Pearson Correlation\\ 
Similarity & & &  &  & \\ \hline 
Co-reference Resolution & 20 (n=1) & 2012 & 590 Notes & 388 Notes & F1  \\ \hline 
Syntactic Parsing & NA & 2016 & NA & NA & F1, UAS, LAS\footnote{UAS: Unlabeled Attachment Score, LAS: Labeled Attachment Score } \\\hline 
Sentence  & NA & 2011-2021 & 518 Notes & 100 Notes & F1 \\
Classification & & & & & \\ 
 \hline 
Document  & 29.33 (n=3) & 2008-2020 & 202-11k Notes & 86-8k Notes& F1, Inversed   \\ 
Classification &  & & & &   Normalized Macro-avg \\ 
 &  & & & &   Mean Absolute  \\ 
 &  & & & & Error (MAE)\\ 
\hline 
Others (Text2SQL,  & 11.67 (n=2) & 2013-2020 & Text2SQL: 37k & Text2SQL: 4k & Error Rate Percentage,  \\
 Speech Recognition)  & & & Records & Records & Acc. \\ 
  & & & SR: 100 Cases  & SR: 100 Cases & \\ \hline  
\end{longtable}

\end{small}

\section*{DISCUSSION}

\subsection*{Lessons Learned from Past Community Interests and Efforts} 

Our scoping review identified a total of 35 papers spanning multiple NLP tasks across both clinical informatics and general domain NLP communities. Among the oldest and most frequent tasks across both communities were NER tasks, which require systems built at the lexicon level. A shift from lexicon level tasks (NER, EL, etc) to document level tasks (MRC, QA, Summ,etc) was observed across the years with a growing interest for language understanding and text generation problems. As the NLP field continues to evolve since the introduction of transformers~\cite{vaswani2017attention} and capacity to build large pre-trained neural language systems~\cite{devlin2019bert,beltagy-etal-2019-scibert}, the breadth of tasks are expected to grow.  In more recent years, the general domain NLP field has contributed natural language understanding and generation tasks~\cite{Radford2018ImprovingLU,roberts2020much,zhang2020pegasus}, but the CI domain remains largely focused in NER~\cite{yetisgen2017automatic,stubbs2017deidentification,shen2021family} and document classification~\cite{filannino2017symptom,stubbs2019cohort}.  This may represent a divergence in interest and lessons learned from the general domain community but can also inform how the clinical NLP domain can expand its scope in tasks and potentially lead to more innovative solutions in clinical applications. 

The first publicly available task appeared in a CI journal by clinical NLP experts working at health systems affiliated with academia~\cite{uzuner2007evaluating}. Several reasons for the earlier appearance by the CI community may include the difficulties in extracting clinical notes and privacy laws protecting patient data for sharing, which requires individuals with direct access to the EHR. The CI community of NLP experts brings together similar computational linguistic knowledge but they collaborate with healthcare providers to tackle the linguistic challenges in EHR data with a better understanding of the medical terms and clinical problems. This is also reflected in the longer history of shared tasks with a focus into a clinical problem (e.g., deidentification~\cite{stubbs2015a,uzuner2007evaluating,stubbs2017deidentification}, clinical trial recruitment~\cite{stubbs2019cohort}, etc). Tasks organizers from the general domain NLP community focused more on fundamental tasks like Parsing~\cite{savkov2016annotating,klassen2016annotating}, NLI~\cite{romanov2018lessons,abacha2019overview}. Most researchers in the general domain NLP community derived from a non-clinical computational background with different motivations to develop cutting-edge technologies with less exposure to the clinical needs of health systems. Majority of papers from the general domain NLP community were not directly associated with clinical applications. Rather, the motivations were to build a foundation for better clinical text understanding, and ultimately facilitate model development for downstream tasks and clinical applications. To further advance the field of clinical NLP, both communities may benefit from a foundational NLP task directed by computational linguistic experts underpinning a major clinical challenge to ultimately improve clinical decision support or health outcomes. 

Another major gap identified from this scoping review were the limitations in generalizability of the data sources.  All the data are relatively homogeneous, deriving from mainly large, urban tertiary academic centers and mainly from single centers with a biased representation of the US population. Further, the notes derived from academic centers also contain a large proportion of notes in the EHR documented by trainees and may not be representative of community hospitals and health systems without trainees and the additional note types derived from them.  The current environment with HIPAA privacy laws and resources for an Enterprise Data Warehouse largely limit the availability of data to centers. Currently, centers with informatics and computational expertise and resources support the data needed for public tasks and these remain limited to well-resourced academic centers. Moving forward, a concerted effort should be placed into sourcing data across geographic regions and hospital types that serve a diversity of patients and encompass various documentation practices. 

\subsection*{Selection of Notes and Benchmarks} 

The discharge summary is the most frequent note type and typically the most detailed about the hospital events and final diagnoses and treatments provided.  While these may be useful for accomplishing certain NLP tasks, their clinical application in real-time remain limited.  Augmented intelligence via clinical decision supports systems frequently ingest data as events happen or use note types with time-sensitive appearance or repeated measures. Discharge summaries are typically the last documentation to resolve what happened during a hospital stay and may not be useful for augmented decision making. Other note types such as radiology and emergency department notes that are time-sensitive or daily progress notes that track disease and treatment plans each day are potentially more useful for real-time NLP applications, which is a goal for many researchers in the field.

F1 scores and overall accuracy are the most frequently used evaluation metrics, but they are only one component in reliability and validity testing.  The extent to which a system measures what it is intended to measure requires multiple validity metrics. Criterion validity metrics with accuracy and correlation scores against reference standards are the de facto standard in tasks. However, construct and content validity are also important. Construct validity is needed when no universally accepted criterion exists to support the concept (or construct) being measured. This may require human evaluation to provide more than just frequentist statistics and better report benchmarks for natural language understanding and generation tasks. Content validity (or face validity), the extent to which the lexicons identified by the system are representative of the concepts the system seeks to measure require more sophisticated approaches that can evaluate semantics and word order like the BERTScore~\cite{zhang2019bertscore}.  In the clinical domain, meeting all the validity metrics may still not be enough. Pragmatic testing through clinical applications with practice simulations that examine the system's effectiveness should also be considered in future tasks. 

\subsection*{Issues in Task/Data Preparation} 
Introducing, preparing and releasing data for a new task requires complex thoughts and actions, yet details on data preparation are often neglected. From the included papers, we identify some issues above that may help to improve future task presentation. Additional issues were inconsistencies reporting data split, and a small amount of papers presented results from pre-trained models without explaining the training set which hinders reproducibility. We also found that the data split sizes reported for most papers did not match with the annotation units. The inconsistency between annotation units and data split unit obfuscates the real problem size. Finally, none of the papers reported how they determined the minimum size of annotations needed to adequately train a model. Recall that even within the same type of tasks, the data size could range substantially from hundreds to thousands (e.g. DocClass). Although it is widely known that annotations are limited to the resources (time, budget, etc), not knowing the minimum sample size raises a crucial question about result reliability issue: will the model performance trained on this dataset be robust enough and trust-worthy? Models developed for tasks like NLI, MRC and QA are data hungry and should be determined a priori, as these tasks require deeper understanding in semantics and relations. We believe by addressing these issues, researchers could make stronger contributions to Clinical NLP. 

\section*{CONCLUSION}


The interests in introducing and participating in clinical NLP tasks are growing with more tasks surfacing each year. The breadth of tasks is also growing with topics varying from tasks with specific clinical applications to those facilitating clinical language understanding and reasoning. It is no doubt that the field will continue to grow and attract more researchers from both general NLP domain and the clinical informatics community. We encourage future work on proposing tasks/shared tasks to overcome barriers in community collaboration, reporting transparency, and consistency of data preparation. 

\subparagraph{ACKNOWLEDGING}
We thank David Bloom (UW-Madison) and Anne Glorioso (UW-Madison) for their review of the search query as computer science librarians. 
 
\makeatletter
\renewcommand{\@biblabel}[1]{\hfill #1.}
\makeatother

\bibliographystyle{vancouver}
\bibliography{main}

\end{document}